\documentclass[12pt,draftcls,onecolumn]{IEEEtran}

\usepackage{hyperref}
\usepackage{makeidx}         
\usepackage{graphicx}        
                                       
\usepackage[pdftex]{color}
\usepackage{multicol}       
\usepackage[bottom]{footmisc}

\usepackage{amsfonts,amsmath,euscript,mathbbol,mathtools,dsfont,mathtools,amssymb, txfonts}

\usepackage{subfig}

\newtheorem{theorem}{\bf Theorem}[section]

\newtheorem{algorithm}{\bf Algorithm}[section]

\newcommand{\bmat}{\left[ \begin{matrix}}
\newcommand{\emat}{\end{matrix} \right]}

\newcommand{\R}{\mathbb R}

\newcommand{\T}{\mathbb T}

\newcommand{\cA}{\mathcal A}
\newcommand{\cB}{\mathcal B}
\newcommand{\cC}{\mathcal C}

\newcommand{\cH}{\mathcal H}
\newcommand{\cI}{\mathcal I}
\newcommand{\cK}{\mathcal K}

\newcommand{\cO}{\mathcal O}
\newcommand{\cP}{\mathcal P}

\newcommand{\cS}{\mathcal S}

\newcommand{\cX}{\mathcal X}

\newcommand{\bh}{\mathbf  h}

\newcommand{\bm}{\mathbf  m}

\newcommand{\bv}{\mathbf  v}

\newcommand{\bx}{\mathbf  x}
\newcommand{\by}{\mathbf  y}

\newcommand{\bA}{\mathbf A}
\newcommand{\bB}{\mathbf B}
\newcommand{\bC}{\mathbf C}

\newcommand{\bJ}{\mathbf J}

\newcommand{\bP}{\mathbf P}

\newcommand{\bX}{\mathbf X}
\newcommand{\bY}{\mathbf Y}

\newcommand{\bSigma}{\boldsymbol{\Sigma}}
\newcommand{\bmu}{\boldsymbol{\mu}}
\newcommand{\bnu}{\boldsymbol{\nu}}

\definecolor{darkgreen}{rgb}{0.0, 0.7, 0.0}

\graphicspath{{./figures/}}

\title{\LARGE \bf
On the Geometry of Message Passing Algorithms for Gaussian Reciprocal Processes
}

\author{Francesca Paola Carli% <-this % stops a space
\thanks{Francesca Paola Carli is with the Department of Electrical Engineering and Computer Science, University of Li\`{e}ge, Belgium, 
and with the Department of Engineering, University of Cambridge, United Kingdom, 
        {\tt\small fpc23@cam.ac.uk }}%
}

\begin{document}

\maketitle
\thispagestyle{empty}
\pagestyle{empty}

%%%%%%%%%%%%%%%%%%%%%%%%%%%%%%%%%%%%%%%%%%%%%%%%%%%%%%%%%%%%%%%%%%%%%%%%%%%%%%%%
\begin{abstract}
Reciprocal processes are acausal generalizations of Markov processes introduced  by Bernstein in 1932.
In the literature, a significant amount of attention has been focused on developing {dynamical models} for reciprocal processes. 
Recently, {probabilistic graphical models} for reciprocal processes have been provided. 
This opens the way to the application of efficient inference algorithms in the machine learning literature to solve the smoothing problem for reciprocal processes. 
Such algorithms are known to converge if the underlying graph is a tree. 
This is not the case for a reciprocal process, whose associated graphical model is a single loop network. 
{The contribution of this paper is twofold. First, we introduce belief propagation for Gaussian reciprocal processes. 
Second, we establish a link between convergence analysis of belief propagation for {Gaussian} reciprocal processes and 
{stability theory} for {differentially positive systems}. }
\end{abstract}

\section{Introduction}

A $\R^n$--valued discrete-time stochastic process $\bX_{k}$ defined over the interval $\cI = [0,N]$ is said to be reciprocal if for any subinterval $[K,L]\subset \cI$, the process in the interior of $[K,L]$ is conditionally independent of the process in $\cI - [K,L]$ given $\bX_{K}$ and $\bX_{L}$.  
From the definition we have that the class of reciprocal processes is  larger than the class of Markov processes: 
Markov processes are necessarily reciprocal, but the converse is not true \cite{Jamison1970}. 
Moreover multidimensional Markov random fields reduce in one dimension to a reciprocal process, not to a Markov process. 

Reciprocal processes were introduced  by Bernstein \cite{Bernstein1932} in 1932, 
who was influenced by an attempt of Schr{\"o}dinger \cite{Schrodinger1932} at giving a stochastic interpretation of quantum mechanics. 
After their introduction by Bernstein, reciprocal processes have been studied in detail by Jamison \cite{Jamison1970,Jamison1974,Jamison1975}, 
Carmichael, Mass{\'e}, Theodorescu \cite{CarmichaelMasseTheodorescu1982} and Levy, Krener, Frezza \cite{Krener1988, LevyFrezzaKrener1990,KrenerFrezzaLevy1991}.   
For more recent literature on reciprocal processes see \cite{CFPP-2011,CFPP-2013}, \cite{CarravettaWhite2012, WhiteCarravetta2011} and references therein. 
As observed in \cite{LevyFrezzaKrener1990} the steady-state distribution of the temperature along a heated ring or a beam subjected to random loads along its length can be modeled in terms of reciprocal processes. 
Relevance for applications is also attested in \cite{CastanonLevyWillsky1985, SrinivasanEdenWillskyBrown2006,PicciCarli2008} where applications to 
tracking of a ship-trajectory \cite{CastanonLevyWillsky1985}, estimation of arm movements \cite{SrinivasanEdenWillskyBrown2006}, and synthesis of textured images \cite{PicciCarli2008} are considered. 

Starting with Krener's work \cite{Krener1988}, a significant amount of attention has been focused on developing \emph{state--space models} for reciprocal processes. 
A second order state--space model for discrete--time \emph{Gaussian} reciprocal processes has been provided in  \cite{LevyFrezzaKrener1990}.  
Modeling in the \emph{finite state space} case has been analyzed separately in  \cite{CarravettaWhite2012} (see also \cite{Carravetta2008}). 

Recently \cite{Carli2016}, \emph{probabilistic graphical models} for reciprocal processes have been provided, which are distribution--independent. 
This opens the way to the application of efficient inference algorithms in the machine learning literature (the belief propagation, a.k.a. sum--product algorithm) to solve the smoothing problem for reciprocal processes. 
Such algorithms are known to converge if the underlying graph is a tree. 
This is not the case for a reciprocal process, whose associated graphical model is a single loop network. 
In \cite{Carli2016} it has been shown that, for the case of \emph{finite--state} reciprocal processes, convergence of the belief propagation iteration boils down to the study of asymptotic stability of a linear time invariant \emph{positive}  system, that can be analyzed via the Hilbert metric. 
This approach is geometric in nature, in that it applies to general linear positive transformations in an arbitrary linear space which map a quite general cone into itself. 
In a recent paper \cite{ForniSepulchre2016}, a generalization of linear positivity, \emph{differential positivity}, has been introduced. 
{Differential positivity} extends linear positivity to the \emph{nonlinear} setting and, similarly to the latter, restricts the asymptotic behavior of a nonlinear system, a result that is proved by exploiting contraction property of differentially positive systems with respect to the Hilbert metric. 
The contribution of this paper is twofold. First, we introduce belief propagation for Gaussian reciprocal processes. 
Second, we 
establish a link between convergence analysis of belief propagation for \emph{Gaussian} reciprocal processes, whose underlying iteration is  nonlinear on the cone of positive definite matrices,
and {stability theory} of  differentially positive systems. 

The paper is organized as follows. 
In Section \ref{sec:Hilbert_metric}, the Hilbert metric is introduced. 
In Section \ref{sec:positive-diff-positive-systems} we briefly touch upon positive and differentially positive systems and on how the property restricts the asymptotic behavior as a consequence of the contraction of the Hilbert metric. 
Reciprocal processes and the associated graphical model are reviewed in Section \ref{sec:rec-proc}. 
In Section \ref{sec:smoothing-RP-via-BP} the belief propagation algorithm is introduced as well as its  specialization for a hidden reciprocal {model}. 
A link between convergence analysis of belief propagation for \emph{Gaussian} reciprocal processes 
and stability theory for differentially positive systems is established in Section \ref{subsec:GaBP}. 
Section \ref{sec:conclusions} ends the paper.

\section{Hilbert metric}\label{sec:Hilbert_metric}

The Hilbert metric was introduced in \cite{Hilbert1895} and is defined as follows.  
Let $\cB$ be a real Banach space and let $\cK$ be a closed solid cone in $\cB$  that is a closed subset $\cK$ with the properties that 
(i) the interior of $\cK$, $\cK^+$, 
is non--empty;  
(ii) $\cK + \cK \subseteq \cK$; 
(iii) $\cK \cap - \cK = \left\{0\right\}$;  
(iv) $\lambda \cK \subset \cK$ for all $\lambda \geq 0$.  
Define the partial order
$$x\preceq y \Leftrightarrow y-x\in \cK\,,$$
and for $x, y \in \cK \backslash \left\{0\right\}$, let 
\begin{align*}
M(x, y) &:= \inf \left\{ \lambda | x -  \lambda y \preceq 0\right\}\\
m(x, y) &:= \sup \left\{\lambda |  x -  \lambda y \succeq 0 \right\}
\end{align*}
The Hilbert metric $d_{\cH}(\cdot, \cdot)$ induced by $\cK$ is defined by 
\begin{equation}\label{eqn:Hilbert_metric}
d_{\cH}\left(x,y\right) := \log\left(\frac{M(x,y)}{m(x,y)}\right), \; \;  x,y \in \cK \backslash \left\{0\right\}\,.
\end{equation}
For example, if $\cB = \R^n$ and the cone $\cK$ is the positive orthant, $\cK = \cO:=  \left\{ (x_1, \dots, x_n)\,:\, x_i \geq 0, \, 1 \leq i \leq n \right\}$, 
then $M(\bx,\by)=\max_{i}(x_i/y_j)$ and $m(\bx,\by)=\min_i(x_i/y_i)$ and the Hilbert metric can be expressed as 
$$
d_{\cH}(\bx,\by) = \log \frac{\max_i(x_i/y_i)}{\min_i{(x_i/y_i)}} \,. 
$$
On the other hand, if $\cB =\cS := \left\{ \bX = \bX^\top \in \R^{n \times n}\right\}$ is the set of symmetric matrices and $\cK = \cP:= \left\{ \bX \succeq 0 \mid \bX \in \cS\right\}$ is the cone of positive semidefinite matrices, then for $\bX,\bY \succ 0$, $M(\bX,\bY)=\lambda_{max}\left(\bX\bY^{-1}\right)$ and $m(\bX,\bY)= \lambda_{min}\left(\bX \bY^{-1}\right)$. Hence the Hilbert metric is 
$$
d_{\cH}(\bX,\bY) = \log \frac{\lambda_{max}\left(\bX \bY^{-1}\right)}{\lambda_{min}\left(\bX \bY^{-1}\right)}\,. 
$$ 

An important property of the Hilbert metric is the following.
The Hilbert metric is a \emph{projective metric} on $\cK$ i.e. it is nonnegative, symmetric, it satisfies the triangle inequality and is such that, for every $x,y \in \cK$, 
$d_{\cH}(x,y)=0$ if and only if $x=\lambda y$ for  some $\lambda > 0$. It follows easily that $d_{\cH}(x,y)$ is constant on rays, that is 
\begin{equation}\label{eqn:Hilbert_metric_invariant_under_scaling}
d_{\cH}\left(\lambda x, \mu y\right) = d_{\cH}\left(x, y\right) \quad \text{for } \lambda, \mu > 0 \,. 
\end{equation}

A second relevant property is in connection with positive operators. 
In \cite{Birkhoff1957} (see also \cite{Bushell1973}) it has been shown that  linear positive operators contract the Hilbert metric. 
This can be used to provide a geometric proof of the Perron--Frobenius theory and, in turn, to prove attractiveness properties of linear positive systems.  
Such a framework, has been recently extended  to prove attractiveness properties of a generalization of linear positive systems, differentially positive systems \cite{ForniSepulchre2016}. 
A brief overview of this theory is the object of the next Section.

\section{Positive and differentially positive systems}\label{sec:positive-diff-positive-systems}

A linear operator $A$ is positive if it maps a cone $\cK$ into itself, i.e. $\bA\cK \subset \cK$ \cite{Bushell1973}. 
For linear dynamical systems $\bx(k+1) = \bA \bx(k)$, $\bA: \R^{n} \rightarrow \R^{n}$, 
\emph{positivity} has the natural interpretation of invariance (and contraction, if the positivity is strict) of the cone $\cK$ along the trajectories of the system. 
Positivity  significantly restricts the behavior of a linear system, as established by Perron--Frobenius theory. 
Under irreducibility assumption, \emph{classical Perron--Frobenius theory} 
guarantees  the existence of a dominant (largest) real eigenvalue for $\bA$ whose associated eigenvector, the Perron-Frobenius vector $\bv_{f}$, is the unique eigenvector that belongs to the interior of $\cK$. 
As a consequence, the subspace spanned by $\bv_{f}$ is an attractor for the linear system, that is, for any vector $\bx \in \cK$, $\bx \neq 0$
\begin{equation}
\lim_{n \rightarrow \infty} \frac{\bA^{n} \bx}{|\bA^{n} \bx|} = \bv_{f}\,. 
\end{equation}
A \emph{geometric interpretation} of Perron--Frobenius theorem has been provided in \cite{Birkhoff1957} (see also \cite{Bushell1973}) where 
existence of a fixed point of the projective space for a strictly positive linear map has been proved as a consequence of contraction properties of the Hilbert metric under the action of a strictly positive linear operator.  
As such, the Perron--Frobenius theorem can be seen as a special case of the contraction mapping theorem. 
Positivity is at the core of a number of properties of Markov {chains}, consensus algorithms and large-scale control.  

\emph{Differential positivity} \cite{ForniSepulchre2016} extends linear positivity to the \emph{nonlinear} setting. 
A nonlinear system  $\bx(k+1) = f(\bx)$ is differentially positive if its linearization along any given trajectory is positive. 
By generalizing {the above--mentioned geometric interpretation of} the Perron--Frobenius theory to a differential framework, it has been shown \cite{ForniSepulchre2016} that differential positivity restricts the asymptotic behavior of a system. 
{Once again, this is a consequence of contraction properties of differentially positive mappings with respect to the Hilbert metric. }
The conceptual picture is that of a cone attached to every point of the state space, defining a cone filed. Contraction of the cone field along the flow eventually constraints the behavior to be one--dimensional. 
The role of the Perron-Frobenius vector in the linear case is played by the Perron-Frobenius vector field, that is an attractor for the linearized dynamic. 
Differentially positive systems encompass positive and monotone systems as particular cases.  
In particular it has been shown in \cite{ForniSepulchre2016} that differentially positive systems reduce to 
the important class of monotone dynamical systems \cite{Smith2008, HirschSmith2005} 
when the state-space is a linear vector space and when the cone field is constant.
In Section \ref{subsec:GaBP} we will show that the iteration underlying the belief propagation algorithm for Gaussian reciprocal processes is indeed a monotone system,   
whose convergence can be studied leveraging on stability theory of differentially positive systems.

\section{Reciprocal Processes}\label{sec:rec-proc}

In this section, we briefly review the definition of reciprocal process and its description in terms of probabilistic graphical models.  
The smoothing problem for a reciprocal process with cyclic boundary conditions is also introduced. 

Recall that a stochastic process $\bX_{t}$ defined on a time interval $\cI$ is said to be \emph{Markov} if, for any $t_0 \in \cI$, 
the past and the future (with respect to $t_0$) are conditionally independent given $\bX_{t_0}$. 
A process is said to be \emph{reciprocal} if, {for each interval $[t_0, t_1] \subset \cI $}, the process in the interior of $[t_{0},t_{1}]$ and the process in $\cI - [t_{0},t_{1}] $ are conditionally independent given $\bX_{t_0}$ and $\bX_{t_1}$. 
More formally, a $(S, \Sigma)$--valued stochastic process on the interval $\cI$ with underlying probability space $\left(\Omega, \cA,P \right)$ is reciprocal if 
\begin{equation}\label{eqn:CIs-RP}
P(AB \mid \bX_{t_{0}}, \bX_{t_{1}}) = P(A \mid \bX_{t_{0}}, \bX_{t_{1}})P(B \mid \bX_{t_{0}}, \bX_{t_{1}}), 
\end{equation}
$\forall t_{0 } < t_{1}$, $[t_{0},t_{1}] \subset \cI$, where  $A$ is the $\sigma$--field generated by the random variables $\left\{ \bX_r:  r \notin [t_{0},t_{1}] \right\}$ 
and $B$ is the $\sigma$-field generated by $\left\{ \bX_r:   r \in (t_{0},t_{1}) \right\}$. 
From the definition it follows that Markov processes are necessarily reciprocal, while the converse is generally  not true \cite{Jamison1970}.
Moreover, a multidimensional Markov random field reduces in one dimension to a reciprocal process, not to a Markov process.  
 
In this paper, we consider reciprocal processes defined on the discrete circle $\T$ with $N+1$ elements $\left\{ 0, 1, \dots N\right\}$  
(which corresponds to imposing the cyclic boundary conditions $\bX_{-1} = \bX_{N}$,  $\bX_{N+1} = \bX_{0}$ \cite{LevyFrezzaKrener1990, Sand96})
so that the additional conditional independence relations 
\begin{align*}
\bX_{0} & \Perp \left\{ \bX_{2}, \dots, \bX_{N-1} \right\} \mid \left\{\bX_{1}, \bX_{N}\right\}\,, \\
\bX_{N} &\Perp \left\{ \bX_{1}, \dots, \bX_{N-2} \right\} \mid \left\{\bX_{0}, \bX_{N-1}\right\}
\end{align*}
hold. 

\begin{figure}[htbp]
\centering 
{\includegraphics[width=0.3\textwidth]{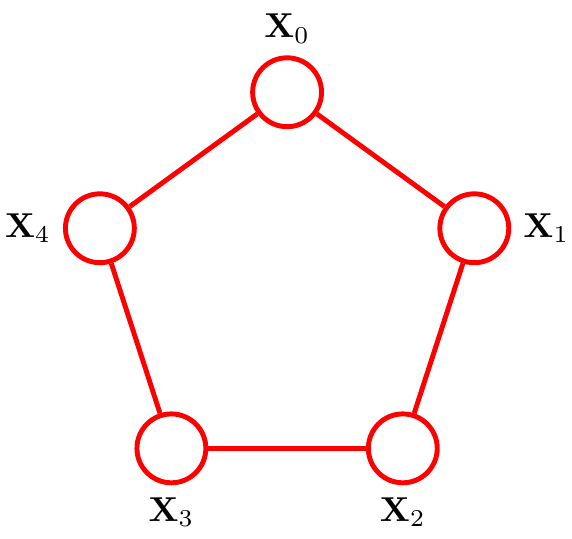}}
\caption{Probabilistic graphical model for a reciprocal process on $\cI=[0,4]$.   
\label{fig:PGM_RP_5nodes}}
\end{figure}

In \cite{Carli2016} it has been shown that the reciprocal process $\left\{ \bX_{k}\right\}$ on $\T$ admits  
a probabilistic graphical model composed of the $N+1$ nodes $\bX_{0}, \bX_{1}, \dots, \bX_{N}$ arranged in a single loop 
undirected graph as shown in Figure \ref{fig:PGM_RP_5nodes}. 

\begin{figure}[htbp]
\centering 
{\includegraphics[width=0.46\textwidth]{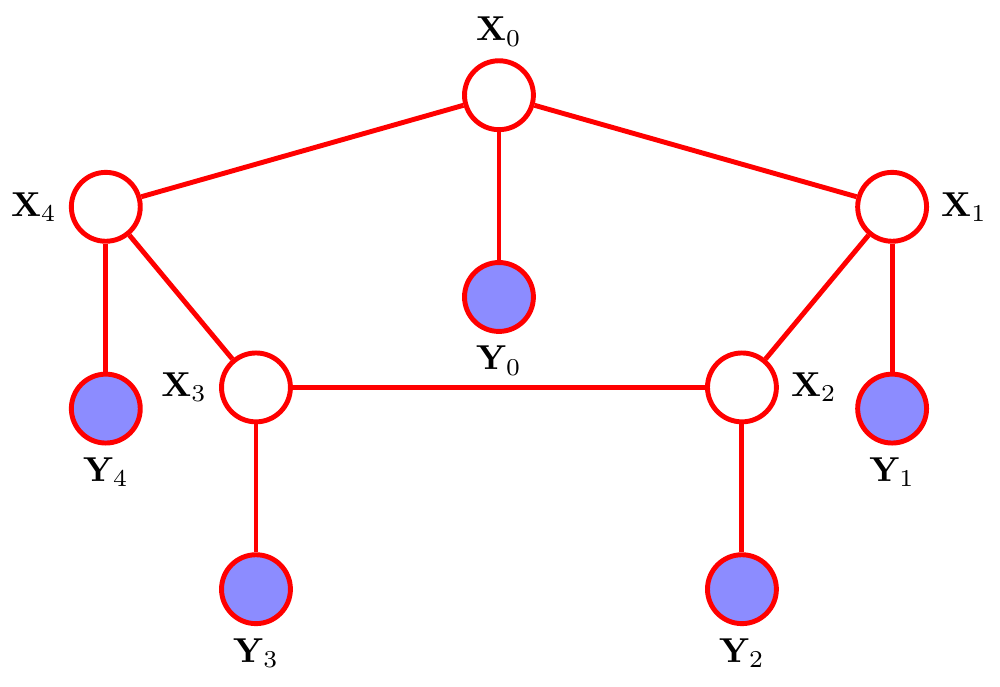}}
\caption{Hidden reciprocal model on $\cI=[0,4]$.  
\label{fig:HRM_5nodes}}
\end{figure}

We now consider a second process $\left\{ \bY_k\right\}$, where, given the state sequence $\left\{ \bX_k\right\}$, the $\left\{\bY_{k}\right\}$  are independent random variables, and for all $k \geq 1$, the conditional probability distribution of $\bY_k$ depends only on $\bX_k$. In applications, $\left\{ \bX_k\right\}$ represents a ``hidden'' process which is not directly observable, 
while the observable process $\left\{ \bY_k\right\}$ represents ``noisy observations'' of the hidden process. 
We shall refer to the pair $\left\{\bX_k, \bY_{k}\right\}$ as a \emph{hidden reciprocal model}. 
The corresponding probabilistic graphical model is illustrated in Figure \ref{fig:HRM_5nodes}.  
The (fixed--interval) \emph{smoothing problem} is to compute, for all $k \in [0,N]$,  the conditional distribution of  $\bX_{k}$ given   
$\bY_{0}, \dots, \bY_{N}$. 
One of the most widespread algorithms for performing inference (solving the smoothing problem) in the graphical models literature is the \emph{belief propagation} algorithm \cite{Pearl1988, KollerFriedman2009,Bishop2006}, 
that will be reviewed in the next Section.

\section{Smoothing of Reciprocal Processes via Belief Propagation} \label{sec:smoothing-RP-via-BP}

In this Section, we first review the belief propagation algorithm \cite{Pearl1988, KollerFriedman2009,Bishop2006} and specialize it for a hidden reciprocal {model}. 
The particular form that the iteration takes  for Gaussian reciprocal processes is discussed in Section \ref{subsec:GaBP}. 

\subsection{Belief Propagation (a.k.a. sum--product) algorithm}

Let $\cH = (E,V)$ be an undirected graphical model over the variables $\left\{ \bX_{0}, \dots, \bX_{N}\right\}$, $\bX_{i} \in  \cX$, $i=0, \dots, N$. 
From the theory of probabilistic graphical models, we have that the joint distribution associated with $\cH$ can be factored as 
\begin{equation}\label{eqn:factorization_UGM_2}
p(\bx) = \frac{1}{Z} \prod_{C \in \cC} \psi_C(\bx_C)\,,
\end{equation}
where $\cC$ denotes a set of maximal cliques in the graph. 
In the following, we will be interested in pairwise Markov random fields -- i.e. a Markov random field in which the joint probability factorizes into a product of bivariate potentials (potentials involving only two variables) -- where each unobserved node $\bX_{i}$ has an associated observed node $\bY_{i}$.  Factorization \eqref{eqn:factorization_UGM_2} then becomes 
\begin{equation}\label{eqn:factorization-pairwise-MRF}
p(\bx_{0:N}, \by_{0:N}) = \prod_{(i,j) \in E} \psi_{ij}(\bx_{i},\bx_{j}) \prod_{i} \psi_{i}(\bx_{i},\by_{i})\,,
\end{equation}
where the $\psi_{ij}(\bx_{i},\bx_{j})$'s are often referred to as the \emph{edge potentials} and the $\psi_{i}(\bx_{i},\by_{i})$'s are often referred to as the \emph{node potentials}. 
The problem we are interested in is finding  marginals of the type  $p(\bx_{i}, \by_{0:N})$ for some hidden variable $\bX_{i}$. 

\begin{figure}
\begin{center}{\includegraphics[width=0.3\textwidth]{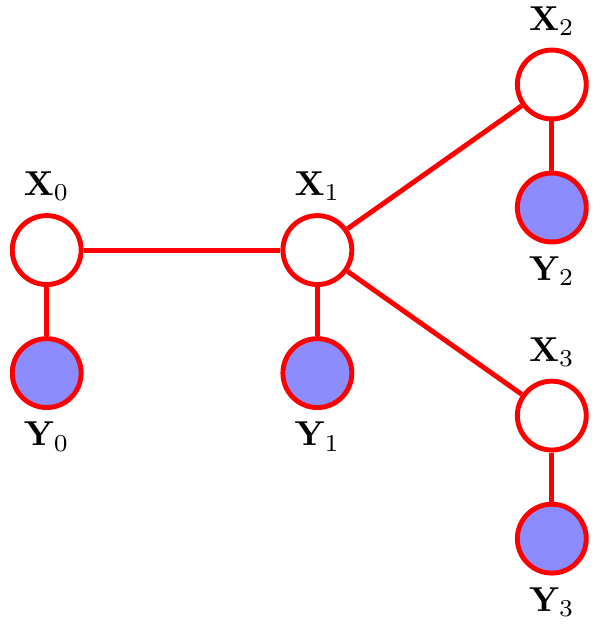}}
\caption{An example of graphical model with four unobserved nodes $\bX_{0}, \dots , \bX_{3}$ and four observed nodes $\bY_{0}, \dots, \bY_{3}$.  \label{fig:Vgraph}}
\end{center}
\end{figure}

The basic idea behind belief propagation is to exploit the factorization properties of the distribution to allow efficient computation of the marginals.  
To fix ideas, consider the graph in Figure \ref{fig:Vgraph} and suppose we want to compute the conditional marginal $p(\bx_{0}\mid \by_{0:3})$.  
A naive application of the definition would suggest that $p(\bx_{0}\mid \by_{0:3})$ can be obtained by summing the joint distribution over all variables except $\bX_{0}$ and then normalize 
\begin{equation}\label{eqn:brute-force_marginal_example}
p(\bx_{0} \mid \by_{0:3})\propto  \int_{\bx_1} \int_{\bx_2} \int_{\bx_3} p(\bx,\by) d\bx_{1} d\bx_{2} d\bx_{3}\,. 
\end{equation}
Nevertheless notice that the joint distribution can be factored as: 
\begin{align}\label{eqn:joint_factored_pdf_example}
p(\bx_{0:3},\by_{0:3}) =  \psi_{0}(\bx_{0}) \psi_{01}(\bx_{0},&\bx_{1}) \psi_{1}(\bx_{1})  \psi_{12}(\bx_{1},\bx_{2}) \nonumber \\
&\psi_{2}(\bx_{2}) \psi_{13}(\bx_{1},\bx_{3}) \psi_{3}(\bx_{3})  \,. 
\end{align}
By plugging in factorization \eqref{eqn:joint_factored_pdf_example} into equation \eqref{eqn:brute-force_marginal_example} and  interchanging the summations and products order, we obtain 
\begin{align}\label{eqn:marginal_x1_example}
p(\bx_{0}\mid\by_{0:3}) &\propto \psi_{0}(\bx_{0})\Bigg[  \int_{\bx_{1}} \psi_{01}(\bx_{0},\bx_{1})\psi_{1}(\bx_{1}) \nonumber \\  &\int_{\bx_{2}} \psi_{12}(\bx_{1},\bx_{2})\psi_{2}(\bx_{2})  
\int_{\bx_{3}} \psi_{13}(\bx_{1},\bx_{3}) \psi_{3}(\bx_{3})  \Bigg]\,. 
\end{align}
This forms the basis for the message--passing algorithm. 

\begin{algorithm}[Belief propagation]
Let $\bX_i$ and $\bX_j$ be two neighboring nodes in the graph. 
We denote by $m_{ij}$ the message that node $\bX_i$ sends to node $\bX_j$, by $m_{ii}$ the message that $\bY_i$ sends to $\bX_i$, 
and by $b_i$ the belief at node $\bX_i$. 
The belief propagation algorithm is as follows: 
\begin{subequations}
\begin{align}
m_{ij} (\bx_{j})  &= \alpha \int_{\bx_{i}} \psi_{ij}(\bx_{i},\bx_{j}) m_{ii}(\bx_{i}) \prod_{k\in \partial i \backslash j} m_{ki} (\bx_{i})\label{eqn:message-update-scalar}\\
b_{i}(\bx_{i}) & = \beta \,\, \, m_{ii}(\bx_{i}) \prod_{k \in \partial i } m_{ki} (\bx_{i}) \label{eqn:beliefs-computation-scalar}
\end{align}
\end{subequations}
{where $\partial i$ denotes the set of {neighbors} of node $\bX_{i}$ and  $\alpha$ and $\beta$ are  normalization constants. }
\end{algorithm}
For example, if one considers \eqref{eqn:marginal_x1_example},  
by setting $m_{ii}(\bx_{i}):=  \psi_{i}(\bx_{i})$ and applying definition \eqref{eqn:message-update-scalar} for the messages, \eqref{eqn:marginal_x1_example} becomes 
\begin{align*}
p(\bx_{0}\hspace{-1mm}\mid \by_{0:3})  & = m_{00}(\bx_{0}) \Big\{  \int_{\bx_{1}} \hspace{-1mm} \psi_{01}(\bx_{0},\bx_{1})\big[ m_{11}(\bx_{1}) \cdot  
  m_{21}(\bx_{1}) \cdot m_{31}(\bx_{1})\big] \Big\} \nonumber \\
& = m_{00}(\bx_{0}) \cdot m_{10}(\bx_{0}) 
\end{align*}
which is of the form \eqref{eqn:beliefs-computation-scalar}, where the marginal $p(\bx_{0},\by_{0:3})$ is computed as the product of incoming messages in the node $\bX_{0}$. 

Observed nodes do not receive messages, and they always transmit the same vector.  
The normalization of messages in equation \eqref{eqn:message-update-scalar} is not theoretically necessary (whether the messages are normalized or not, the beliefs $b_{i}$ will be identical)
but helps avoiding numerical underflow problems and improving numerical stability of the algorithm.  
Finally, notice that equation \eqref{eqn:message-update-scalar} does not specify the order in which the messages are updated. 
In this paper we assume that all nodes simultaneously update their messages in parallel. 
This naturally leads to \emph{loopy belief propagation}, where the update rule \eqref{eqn:message-update-scalar} is applied to graphs that are not a tree (like the single loop network associated to a reciprocal process).

\subsection{Belief Propagation for general (non necessarily Gaussian) Hidden Reciprocal {Models}}

If the considered graph is the single--loop hidden reciprocal {model} in Figure \ref{fig:HRM_5nodes}, 
expressions \eqref{eqn:message-update-scalar} and  \eqref{eqn:beliefs-computation-scalar} for the message and belief updates simplify, each node having only two neighbors.  
Moreover we can distinguish between two classes of messages, one propagating in the direction of increasing indexes (clockwise) 
and one propagating in the direction of decreasing indexes (anticlockwise) in the loop.  
The overall algorithm with parallel scheduling policy is as follows: 

\begin{algorithm}\textbf{[(Parallel) belief propagation algorithm for a hidden reciprocal {model}]} \label{alg:BP-hidden-reciprocal-chain}
\begin{enumerate}
	\item Initialize all messages $\bm_{ij}^{(0)}$ to some initial value $\bar{\bm}_{ij}^{(0)}$. 

	\item Iteratively apply the updates
	\begin{subequations}\label{eqn:forward-backward-messages_on-a-chain}
\begin{align}
m_{k-1,\,k}^{(t+1)} (\bx_{k})  &=\alpha_{f}\hspace{-1mm}  \int_{\bx_{k-1}} \hspace{-3mm}\psi_{k-1,k}(\bx_{k-1},\bx_{k}) m_{k-1,k-1}(\bx_{k-1})  
 m_{k-2,k-1}^{(t)} (\bx_{k-1}) \label{eqn:mess_chain_forward} \\
m_{k+1,\,k}^{(t+1)} (\bx_{k})  &= \alpha_{b}\hspace{-1mm} \int_{\bx_{k+1}}  \hspace{-3mm} \psi_{k+1,k}(\bx_{k+1},\bx_{k}) m_{k+1,k+1}(\bx_{k+1}) 
 m_{k+2,k+1}^{(t)} (\bx_{k+1})\,.  \label{eqn:mess_chain_backward} 
\end{align}
\end{subequations}
	
	\item For each $\bX_{i}$, $i=0, \dots, N$ compute the marginals 
	\begin{equation}\label{eqn:beliefs-computation-matricial-parallel}
	b_{k}(\bx_{k})   = \beta \,\, \, m_{kk}(\bx_{k}) \left[ m_{k-1,k}^{(t_{max})} (\bx_{k}) \cdot  m_{k+1,k}^{(t_{max})} (\bx_{k}) \right]\,. 
	\end{equation}
	
\end{enumerate}
\end{algorithm}
For tree-structured graphs, when $t_{max}$ is larger than the diameter of the tree (the length of longest shortest path  between any two vertices of the graph), the algorithm converges to the correct marginal. 
{Convergence analysis of belief propagation for a single--loop network like the one associated to a reciprocal process has been carried out in \cite{Weiss2000, WeissFreeman2001}, 
{where the finite state space case and the case of Gaussian distributed random variables have been separately analyzed.  }
{For Gaussian distributed random variables it has been shown that the belief propagation algorithm converges to the correct mean, and formulas that link the correct covariance and the estimated one have been provided. }
Intrigued by the similarities observed in \cite{WeissFreeman2001} between convergence of finite--state and Gaussian belief propagation on a single loop network (``Although
there are many special properties of gaussians, we are struck by the similarity of the analytical results reported here for gaussians and
the analytical results for single loop and general distributions reported in  \cite{Weiss2000}''), 
that in the former case has been shown to be linked to contraction properties of the Hilbert metric \cite{Carli2016}, 
in Section \ref{subsec:GaBP} we revisit convergence analysis for Gaussian belief propagation in the single--loop network and 
establish a link with {stability theory} of differentially positive systems, which is also rooted in contraction properties of the Hilbert metric.
}

\section{Gaussian Belief Propagation for a Hidden Reciprocal {Model}}\label{subsec:GaBP}

For Gaussian distributed variables, messages and beliefs are Gaussians and the belief propagation updates can be written explicitly in terms of means and covariances. 
In other words, iterations \eqref{eqn:mess_chain_forward}, \eqref{eqn:mess_chain_backward} on the infinite dimensional space of nonnegative measurable functions 
become iterations on the finite dimensional spaces (cones) of nonnegative vectors and positive definite matrices. 
By showing that the latter defines a nonlinear monotone system, we establish a connection between convergence analysis of belief propagation 
for Gaussian reciprocal processes and {stability theory} of differentially positive systems.  

To start, notice that, for Gaussian distributed variables, the factorization \eqref{eqn:factorization-pairwise-MRF} becomes 
\begin{align}\label{eqn:factorization-pairwise-Gaussian-MRF}
p(\bx, \by) \propto & \prod_{(i,j) \in E} {\rm exp}\left\{ -\frac12 \bmat \bx_{i} & \bx_{j}\emat \bP_{ij} \bmat \bx_{i} \\ \bx_{j}\emat\right\} \nonumber \\
&\prod_{i \in V} {\rm exp}\left\{ -\frac12 \bmat \bx_{i} & \by_{i}\emat \bP_{ii} \bmat \bx_{i} \\ \by_{i}\emat\right\}
\end{align}
where we assume that the $\bP_{ij}$'s are all positive semidefinite and, together with the $\bP_{ii}$'s,  can be block partitioned as 
$$
\bP_{ij} = \bmat \bP_{ij}(1,1) & \bP_{ij}(1,2)\\
\bP_{ij}(1,2)^{\top}&\bP_{ij}(2,2)\emat \, 
$$
and
$$
\bP_{ii} = \bmat \bP_{ii}(1,1) & \bP_{ii}(1,2)\\
\bP_{ii}(1,2)^{\top}&\bP_{ii}(2,2)\emat\,, 
$$

Denote by $\bJ_{ij}$ ($\bh_{ij}$) the precision matrix  (resp. potential vector) of the message from $\bX_{i}$ to $\bX_{j}$, 
and by $\hat{\bJ}_{ii}$ ($\hat{\bh}_{ii}$) the precision matrix  (resp. potential vector) of the belief (estimated marginal posterior) $b(\bx_{i}) := \hat{p}(\bx_{i} \mid \by)$. 
Also recall that $\bP_{ij}$ represents the precision matrix associated to the edge potential $\psi_{ij}$
 and $\bP_{ii}$ ($\bnu_{ii}$) the precision matrix  (resp. potential vector) of the node potential $\psi_{ii}$. 
By taking into account the expressions of the node and edge potentials in \eqref{eqn:factorization-pairwise-Gaussian-MRF}, 
for Gaussian distributed random variables, messages \eqref{eqn:mess_chain_forward}, traveling clockwise in the loop,  become
\begin{subequations}
	\begin{align}
	\bJ_{k-1,\, k} & = \bP_{k-1, \, k}({2,2})- \bP_{k-1, \, k}(1,2) \Big[  \bP_{k-1, \, k}({1,1}) \nonumber \\ 
	 & \quad +\bP_{k-1, \, k-1}(1,1) +   \bJ_{k-2,\, k-1} \Big]^{-1} \bP_{k-1,\, k}(1,2)^{\top} \label{eqn:GaBP_J_forward}\\
	\bh_{k-1,\, k}  & = - \bP_{k-1, \, k}(1,2) \Big[  \bP_{k-1,\, k}(1,1) + \bP_{k-1, \, k-1}(1,1) \nonumber \\  
	& \quad +   \bJ_{k-2, \, k-1} \Big]^{-1} \left( \bnu_{k-1, \, k-1} +   \bh_{k-2, \, k-1}  \right)  \label{eqn:GaBP_h_forward}
	\end{align}
\end{subequations}
while messages \eqref{eqn:mess_chain_backward},  traveling anticlockwise  in the loop, are given by 
\begin{subequations}
\begin{align}
	\bJ_{k+1,\, k} & = \bP_{k,k+1}({1,1})- \bP_{k, k+1}(1,2) \Big(  \bP_{k,k+1}({2,2})  \nonumber \\  
	& \quad + \bP_{k+1,k+1}(1,1) +  \bJ_{k+2,k+1} \Big)^{-1} \bP_{k, k+1}(1,2)^{\top} \label{eqn:GaBP_J_backward}\\
	\bh_{k+1,\, k} & = - \bP_{k,k+1}(1,2) \Big(  \bP_{k,k+1}(1,1) + \bP_{k+1,k+1}(1,1)  \nonumber \\ 
	& \quad +  \bJ_{k+2,k+1} \Big)^{-1} \left( \bnu_{k+1,k+1} +  \bh_{k+2,k+1}  \right) \,. \label{eqn:GaBP_h_backward}
\end{align}
\end{subequations}
The estimated beliefs  (estimated posterior mean and covariance) at node $\bX_{k}$ are 
\begin{subequations}
	\begin{align}
	\hat{\bJ}_{k} & = \bP_{kk}(1,1) + \bJ_{k-1,k} + \bJ_{k+1,k}  \\
	\hat{\bh}_{k} & = \bnu_{kk} + \bh_{k-1,k} + \bh_{k+1,k}  
	\end{align}
\end{subequations}
from which the estimated mean vector and covariance matrix associated with the posterior marginals are 
\begin{equation}\label{eqn:GaBP-mean-covariance}
\hat{\bmu}_{k} = \hat{\bJ}_{k} \, \hat{\bh}_{k}, \qquad \hat{\bSigma}_{k} = \hat{\bJ}_{k} ^{-1}\,. 
\end{equation}
Equations \eqref{eqn:GaBP_h_forward}, \eqref{eqn:GaBP_h_backward} provide a \emph{linear} time--varying recursive relation for the computation of message potentials vectors, since they express  $\bh_{k-1,\,k}$ ($\bh_{k,k+1}$) as a linear function of the message potential  on the ``previous'' (resp., ``successive'') link. 
On the other hand, both the maps \eqref{eqn:GaBP_J_forward}, \eqref{eqn:GaBP_J_backward} are of the form 
\begin{equation}\label{eqn:GaBP_Jmap}
\psi(\bJ)   = \bA_{k} - \bB_{k} \left(  \bC_{k} +   \bJ \right)^{-1} \bB_{k}^{\top}
\end{equation}
i.e. they provide a \emph{nonlinear} {time--varying} recursive relation for the computation of the message precision matrix $\bJ_{k-1,k}$ ($\bJ_{k,k+1}$) as a function of the message precision matrix on the ``previous'' (resp., ``successive'') link in the graph. 

\begin{theorem} \label{thm:onestep-Jmap--monotone}
Suppose that $\bA_{k}, \, \bC_{k} \, \in \cS$ (set of symmetric matrices) and that $\bC_{k} + \bJ$ is invertible. 
The map \eqref{eqn:GaBP_Jmap} is monotone (describes a monotone dynamical system).  
\end{theorem}

\begin{IEEEproof} 
The map \eqref{eqn:GaBP_Jmap} is the composition of the following transformations: 
(i) $\tau_{A}(\bJ) = \bJ + \bA$, 
(ii) $\tau_{C}(\bJ) = \bJ+\bC$, 
(iii) $\gamma_{B}(\bJ) = \bB \bJ \bB^\top$, 
(iv) $\sigma(\bJ)=\bJ^{-1}$, and 
(v) $\rho(\bJ) = - \bJ$ 
defined on $\cP^{+}$.  In fact we have 
$$
\psi(\bJ) = ( \tau_{A} \circ \rho \circ  \gamma_{B}  \circ   \sigma  \circ \tau_{C}) (\bJ)\,. 
$$
The transformations $\tau_A$ (equiv. $\tau_C$) and the congruence transformation $\gamma_B$ are order preserving (monotone increasing). 
The inverse map $\sigma$  and the map $\rho$ are order reversing (monotone decreasing). 
Since in the composition there is an even number of order reversing factors, the composite map $\psi$ is order preserving \cite{Schechter2005}. 
\end{IEEEproof}

We now observe the following. 
Without loss of generality, consider the message that $\bX_{N}$ sends to $\bX_{0}$. 
By the update equation \eqref{eqn:mess_chain_forward}, the message that $\bX_{N}$ sends to $\bX_{0}$ at time {$t+N+1$}  
depends on the message that $\bX_{N}$ received from $\bX_{N-1}$ at time {$t+N$}, 
so that, in terms of precision matrices of the messages, we can write  
\begin{equation} \label{eqn:mN0(t+N)}
\bJ_{N 0}^{{(t+N+1)} }= \psi^{f}_{N0} \left( \bJ_{N-1, N}^{(t+N)} \right)
\end{equation} 
where $\psi^{f}_{N0}$ is the nonlinear transformation \eqref{eqn:GaBP_J_forward}. 
Similarly, the message that $\bX_{N-1}$ sends to $\bX_{N}$ at time $t+N$
depends on the message that $\bX_{N-1}$ received from $\bX_{N-2}$  at time $t+N-1$
\begin{equation}\label{eqn:mN-1N(t+N-1)}
\bJ_{N-1 , N}^{(t+N)} = \psi^{f}_{N-1,N} \left( \bJ_{N-2, N-1}^{(t+N-1)} \right)
\end{equation} 
One can continue expressing each message in terms of the one received from the neighbor until we go back in the loop to $\bX_{0}$: 
the message that $\bX_{0}$ sends to $\bX_{1}$ at time $t+1$ is a function of the message that $\bX_{N}$ sent to $\bX_{0}$  at time $t$
\begin{equation}\label{eqn:m0N(t+1)}
\bJ_{0 1}^{(t+1)} = \psi^{f}_{01} \left(  \bJ_{N 0}^{(t)} \right)\,. 
\end{equation}  
By putting together \eqref{eqn:mN0(t+N)}--\eqref{eqn:m0N(t+1)}, 
one gets that the message that $\bX_{N}$ sends to $\bX_{0}$ at a given time step depends on the message that $\bX_{N}$ sent to $\bX_{0}$ $N+1$ time steps ago. 
In particular, if we denote by $\Psi_{N0}^{f}$ the map  
\begin{equation}\label{eqn:CN0}
\Psi_{N0}^{f} = \psi^{f}_{N0} \circ \psi^{f}_{N-1,N}  \circ \dots \circ \psi^{f}_{01}\,,
\end{equation}
the precision matrix of the message that  $\bX_{N}$ sends to $\bX_{0}$ satisfy the recursion 
\begin{equation}\label{eqn:dyn-sys_BP_N0}
\bJ_{N 0}^{(t+N +1)}=\Psi_{N0}^{f} (\bJ_{N 0}^{(t)})\,. 
\end{equation} 
where the map $\Psi_{N0}^{f}$ is given by the composition $\Psi := \psi_{N,0}^{f} \circ \psi_{N-1,N}^{f}  \circ \dots \circ \psi_{0,1}^{f}$, with maps $\psi_{k-1,k}^{f}$ as in \eqref{eqn:GaBP_J_forward}. 
{The map $\Psi_{N0}^{f}$ links the precision matrix of the message on the link $\bX_{N}-\bX_{0}$ to the precision matrix of the message on the same link one loop ago, and it is time--invariant 
(does not vary from the first, to the second, to the third etc. loop) where the time to complete a loop has been taken as the time unit in iteration \eqref{eqn:dyn-sys_BP_N0}. }
Moreover such a map is nonlinear and monotone because composition of monotone maps (by Theorem \ref{thm:onestep-Jmap--monotone}).   
By the discussion in Section \ref{sec:positive-diff-positive-systems} it follows that convergence analysis of Gaussian belief propagation for a hidden reciprocal {model} can be carry out 
leveraging on stability theory of differentially positive systems. 
A detailed analysis is the subject of ongoing work.

\section{Conclusions}\label{sec:conclusions}

In this paper we have introduced belief propagation for performing inference for Gaussian reciprocal processes. 
Intrigued by the similarities observed in \cite{WeissFreeman2001} between convergence results for  finite state space and Gaussian belief propagation on a single loop network, 
that in the finite state space case has been shown to be linked to contraction properties of the Hilbert metric \cite{Carli2016}, 
we have revisited convergence analysis for Gaussian belief propagation in the single--loop network  
establishing a link with stability theory of differentially positive systems, which is also rooted in contraction properties of the Hilbert metric.

%%%%%%%%%%%%%%%%%%%%%%%%%%%%%%%%%%%%%%%%%%%%%%%%%%%%%%%%%%%%%%%%%%%%%%%%%%%%%%%%
%\section{ACKNOWLEDGMENTS}

%The authors gratefully acknowledge the contribution of National Research Organization and reviewers' comments.

%%%%%%%%%%%%%%%%%%%%%%%%%%%%%%%%%%%%%%%%%%%%%%%%%%%%%%%%%%%%%%%%%%%%%%%%%%%%%%%%
\bibliographystyle{plain}
\bibliography{biblio_FilteringHRM-via-MP}

\end{document}